\documentclass{article} 
\usepackage{nips10submit_e,times}
\usepackage{amsmath}
\usepackage{amssymb}
\usepackage{graphicx}
\usepackage{tabularx}
\usepackage{rotating}
\usepackage{algorithm}
\usepackage{algorithmic}
\usepackage{boxedminipage}

\def\tps{^{\mathsf{T}}}

\def\invtps{^{-\mathsf{T}}}
\def\tr{\text{tr}}

\def\Re{{\mathbb R}}
 \DeclareMathOperator*{\argmin}{arg\,min}
  \DeclareMathOperator*{\argmax}{arg\,max}

\title{Predictive State Temporal Difference Learning}

\author{
Byron Boots \\
Machine Learning Department\\
Carnegie Mellon University\\
Pittsburgh, PA 15213 \\
\texttt{beb@cs.cmu.edu} \\
\And
Geoffrey J. Gordon\\
Machine Learning Department\\
Carnegie Mellon University\\
Pittsburgh, PA 15213 \\
\texttt{ggordon@cs.cmu.edu} \\
}

\nipsfinalcopy 

\begin{document}

\maketitle

\begin{abstract}
We propose a new approach to value function approximation which combines linear temporal difference reinforcement learning with subspace identification. In practical applications, reinforcement learning (RL) is complicated by the fact that state is either high-dimensional or partially observable. Therefore, RL methods are designed to work with \emph{features} of state rather than state itself, and the success or failure of learning is often determined by the suitability of the selected features. By comparison, subspace identification (SSID) methods are designed to \emph{select} a feature set which preserves as much information as possible about state. In this paper we connect the two approaches, looking at the problem of reinforcement learning with a large set of features, each of which may only be marginally useful for value function approximation. We introduce a new algorithm for this situation, called Predictive State Temporal Difference (PSTD) learning. As in SSID for predictive state representations, PSTD finds a linear \emph{compression operator} that projects a large set of features down to a small set that preserves the maximum amount of predictive information. As in RL, PSTD then uses a Bellman recursion to estimate a value function. We discuss the connection between PSTD and prior approaches in RL and SSID.  We prove that PSTD is statistically consistent, perform several experiments that illustrate its properties, and demonstrate its potential on a difficult optimal stopping problem.
 \end{abstract}

\section{Introduction and Related Work} 
We examine the problem of estimating a policy's value function within a decision
process in a high dimensional and partially-observable environment, when the parameters of the process are unknown. 
In this situation, a common strategy is to employ a \emph{linear architecture} and represent the value function as a linear combination of \emph{features} of (sequences of) observations. 
A popular family of model-free algorithms called temporal difference (TD) algorithms~\cite{Sutton88} can then be used to estimate the parameters of the value function. Least-squares TD (LSTD) algorithms~\cite{Boyan99, Bradtke96linearleast-squares, Lagoudakis2003} exploit the linearity of the value function to find the optimal parameters in a least-squares sense from time-adjacent samples of features.

Unfortunately, choosing a good set of features is hard. The features
must be predictive of future reward, and the set of features must be
small relative to the amount of training data, or TD learning will be
prone to overfitting. The problem of selecting a small set of
reasonable features has been approached from a number of different
perspectives.  In many domains, the features are selected by hand
according to expert knowledge; however, this task can be difficult and
time consuming in practice. Therefore, a considerable amount of
research has been devoted to the problem of automatically identifying
features that support value function approximation. 

Much of this research is devoted to finding sets of features when the
dynamical system is known, but the state space is large and difficult
to work with. For example, in a large fully observable Markov decision
process (MDP), it is often easier to estimate the value function from
a low dimensional set of features than by using state directly.  So,
several approaches attempt to automatically discover a small set of
features from a given larger description of an MDP, e.g., by using a
spectral analysis of the state-space transition graph to discover a
low-dimensional feature set that preserves the graph
structure~\cite{Mahadevan05a, Mahadevan05b, Johns2007}.

Partially observable Markov decision processes (POMDPs) extend MDPs to
situations where the state is not directly
observable~\cite{astrom65,sondik,cassandra94}. In this circumstance,
an agent can plan using a continuous \emph{belief state} with
dimensionality equal to the number of hidden states in the POMDP\@.
When the number of hidden states is large, dimensionality reduction in
POMDPs can be achieved by projecting a high dimensional belief space
to a lower dimensional one; of course, the difficulty is to find a
projection which preserves decision quality. Strategies for finding
good projections include value-directed compression~\cite{Poupart2002}
and non-negative matrix factorization~\cite{Li2007,
  Theocharous2010}. The resulting model after compression is a
Predictive State Representation (PSR)~\cite{littman02, singh04}, an
Observable Operator Model~\cite{jaeger00a}, or a multiplicity
automaton~\cite{Even-dar05}.  Moving to one of these representations
can often compress a POMDP by a large factor with little or no loss in
accuracy: examples exist with arbitrarily large lossless compression
factors, and in practice, we can often achieve large compression
ratios with little loss.



The drawback of all of the approaches enumerated above is that they
first assume that the dynamical system model is known, and only then
give us a way of finding a compact representation and a value
function. In practice, we would like to be able to find a good set of
features, \emph{without prior knowledge of the system model}. Kolter
and Ng~\cite{Kolter2009} contend with this problem from a sparse
feature selection standpoint. Given a large set of possibly-relevant
features of observations, they proposed augmenting LSTD by applying an
$L_1$ penalty to the coefficients, forcing LSTD to select a sparse set
of features for value function estimation. The resulting algorithm,
LARS-TD, works well in certain situations (for example, see
Section~\ref{sec:RR-POMDP}), but only if our original large set of
features contains a small subset of highly-relevant features.


Recently, Parr et al.\ looked at the problem of value function
estimation from the perspective of both model-free and model-based
reinforcement learning~\cite{Parr2008a}.  The model-free approach
estimates a value function directly from sample trajectories, i.e.,
from sequences of feature vectors of visited states.  The model-based
approach, by contrast, first learns a model and then computes the
value function from the learned model.  Parr et al.\ compared LSTD (a
model-free method) to a model-based method in which we first learn a
linear model by viewing features as a proxy for state (leading to a
linear transition matrix that predicts future features from past
features), and then compute a value function from this approximate
model. Parr et al.\ demonstrated that these two approaches compute
exactly the same value function~\cite{Parr2008a}, formalizing a fact
that has been recognized to some degree before~\cite{Boyan99}.

In the current paper, we build on this insight, while simultaneously
finding a compact set of features using powerful methods from system
identification. First, we look at the problem of improving LSTD from a
model-free predictive-bottleneck perspective: given a large set of
features of history, we devise a new TD method called Predictive State
Temporal Difference (PSTD) learning that estimates the value function
through a bottleneck that preserves only \emph{predictive} information
(Section~\ref{sec:pred}). Intuitively, this approach has some of the
same benefits as LARS-TD: by finding a small set of predictive
features, we avoid overfitting and make learning more
data-efficient. However, our method differs in that we identify a
small \emph{subspace} of features instead of a sparse \emph{subset} of
features. Hence, PSTD and LARS-TD are applicable in different
situations: as we show in our experiments below, PSTD is better when
we have many marginally-relevant features, while LARS-TD is better
when we have a few highly-relevant features hidden among many
irrelevant ones.

Second, we look at the problem of value function estimation from a
model-based perspective (Section~\ref{PSRs}). Instead of learning a
linear transition model from features, as in~\cite{Parr2008a}, we use
subspace identification~\cite{rosencrantz04, Boots2010b} to learn a
PSR from our samples.
Then we compute a value function via the Bellman equations for our
learned PSR\@.  This new approach has a substantial benefit: while the
linear feature-to-feature transition model of~\cite{Parr2008a} does
not seem to have any common uses outside that paper, PSRs have been
proposed numerous times on their own merits (including being invented
independently at least three times), and are a strict generalization
of POMDPs.

Just as Parr et al.\ showed for the two simpler methods, we show that
our two improved methods (model-free and model-based) are equivalent.
This result yields some appealing theoretical benefits: for example,
PSTD features can be explicitly interpreted as a statistically
consistent estimate of the true underlying system state.  And, the
feasibility of finding the true value function can be shown to depend on the
\emph{linear dimension} of the dynamical system, or equivalently, the
dimensionality of the predictive state representation---\emph{not} on
the cardinality of the POMDP state space. Therefore our representation
is naturally ``compressed'' in the sense of~\cite{Poupart2002},
speeding up convergence.

The improved methods also yield practical benefits; we demonstrate
these benefits with several experiments.  First, we compare PSTD to
LSTD and LARS-TD on a synthetic example using different sets of
features to illustrate the strengths and weaknesses of each
algorithm. Next, we apply PSTD to a difficult optimal stopping problem
for pricing high-dimensional financial derivatives. A significant
amount of work has gone into hand tuning features for this problem. We
show that, if we add a large number of weakly relevant features to
these hand-tuned features, PSTD can find a predictive subspace which
performs much better than competing approaches, improving on the best
previously reported result for this particular problem by a
substantial margin.

The theoretical and
empirical results reported here suggest that, for many applications
where LSTD is used to compute a value function, PSTD can be simply
substituted to produce better results.

\section{Value Function Approximation}
We start from a discrete time dynamical system with a set of states
$\mathcal{S}$, a set of actions $A$, a distribution over initial
states $\pi_0$, a state transition function $T$, a reward function
$\mathcal{R}$, and a discount factor $\gamma \in [0,1]$. We seek a
policy $\pi$, a mapping from states to actions. The notion of a value
function is of central importance in reinforcement learning: for a
given policy $\pi$, the value of state $s$ is defined as the expected
discounted sum of rewards obtained when starting in state $s$ and
following policy $\pi$, $J^\pi(s) =
\mathbb{E}\left[\sum_{t=0}^\infty\gamma^t\mathcal{R}(s_t)\mid s_0 = s,
  \pi \right ]$. It is well known that the value function must obey
the Bellman equation
\begin{align}\label{J-def}
J^\pi(s) &= \mathcal{R}(s) + \gamma \sum_{s'}J^\pi(s')\Pr[s' \mid s, \pi(s)]
\end{align}
If we know the transition function $T$, and if the set of states
$\mathcal S$ is sufficiently small, we can use~(\ref{J-def}) directly
to solve for the value function $J^\pi$.  We can then execute the
\emph{greedy policy} for $J^\pi$, setting the action at each state to
maximize the right-hand side of~(\ref{J-def}).

However, we consider instead the harder problem of estimating the
value function when $s$ is a \emph{partially observable} latent
variable, and when the transition function $T$ is unknown. In this
situation, we receive information about $s$ through observations from
a finite set $O$. Our state (i.e., the information which we can use to
make decisions) is not an element of $\mathcal S$ but a \emph{history}
(an ordered sequence of action-observation pairs $h = a_1^ho_1^h
\hdots a_t^ho_t^h$ that have been executed and observed prior to time
$t$).  If we knew the transition model $T$, we could use $h$ to infer
a belief distribution over $\mathcal S$, and use that belief (or a
compression of that belief) as a state instead; below, we will discuss
how to learn a compressed belief state.  Because of partial
observability, we can only hope to predict reward conditioned on
history, $\mathcal R(h) = \mathbb E[\mathcal R(s)\mid h]$, and we must
choose actions as a function of history, $\pi(h)$ instead of $\pi(s)$.

Let $\mathcal{H}$ be the set of all possible histories. $\mathcal{H}$ is often very large or infinite, so
instead of finding a value separately for each history, we focus on
value functions that are linear in \emph{features} of histories
\begin{align}
J^\pi(s) = w\tps \phi^\mathcal{H}(h)
\end{align}
Here $w \in \mathbb{R}^j$ is a parameter vector and
$\phi^\mathcal{H}(h) \in \mathbb{R}^j$ is a feature vector
for a history $h$.
 So, we can
rewrite the Bellman equation as%
\begin{align}
w\tps \phi^\mathcal{H}(h) &= \mathcal{R}(h) + \gamma \sum_{o\in O}
w\tps \phi^\mathcal{H}(h\pi o) \Pr[h\pi o \mid h \pi]
\end{align} 
where $h\pi o$ is history $h$ extended by taking action $\pi(h)$ and observing $o$.

\subsection{Least Squares Temporal Difference Learning}
\label{sec:LSTD}
In general we don't know the transition probabilities $ \Pr[h\pi o\mid
h]$, but we do have samples of state features $\phi^\mathcal{H}_t =
\phi^\mathcal{H}(h_t)$, next-state features $\phi^{\mathcal{H}}_{t+1}
= \phi^\mathcal{H}(h_{t+1})$, and immediate rewards $\mathcal{R}_t = \mathcal{R}(h_t)$. We can thus
\emph{estimate} the Bellman equation
\begin{align}
\label{eq:Bellman}
w\tps\phi^\mathcal{H}_{1:k}  &\approx \mathcal{R}_{1:k} +  \gamma w\tps\phi^{\mathcal{H}}_{2:k+1}
\end{align}
(Here we have used the notation $\phi^{\mathcal H}_{1:k}$ to mean the matrix whose
columns are $\phi^{\mathcal H}_t$ for $t=1\ldots k$.)
We can can immediately attempt to estimate the parameter $w$ by
solving the linear system in the least squares sense: $ \hat w\tps=
\mathcal{R}_{1:k} \left ( \phi^\mathcal{H}_{1:k} -  \gamma \phi^{\mathcal{H}}_{2:k+1}
\right )^\dagger$, where $^\dagger$ indicates the Moore--Penrose
pseudo-inverse. However, this solution is
biased~\cite{Bradtke96linearleast-squares}, since the independent variables
$\phi^\mathcal{H}_t -  \gamma \phi^{\mathcal{H}}_{t+1}$ are \emph{noisy}
samples of the expected difference $\mathbb E[\phi^\mathcal{H}(h) -  \gamma
\sum_{o\in O} \phi^\mathcal{H}(h \pi o) \Pr[h \pi o\mid h]]$. In
other words, estimating the value function parameters $w$ is an
\emph{error-in-variables} problem.

The \emph{least squares temporal difference} (LSTD) algorithm provides
a consistent estimate of the independent variables by right
multiplying the approximate Bellman equation
(Equation~\ref{eq:Bellman}) by ${\phi_t^\mathcal{H}}\tps$. The
quantity ${\phi_t^\mathcal{H}}\tps$ can be viewed as an
\emph{instrumental} variable~\cite{Bradtke96linearleast-squares},
i.e., a measurement that is correlated with the true independent
variables, but uncorrelated with the noise in our estimates of these variables.\footnote{The LSTD algorithm can also be theoretically justified as the result of an application of the Bellman operator followed by an orthogonal projection back onto the row space of  $\phi^\mathcal{H}$~\cite{Lagoudakis2003}.} The value function parameter $w$ may then be estimated as follows:
\begin{align}
\label{eq:LSTD}
\hat w\tps&=  \frac{1}{k}\sum_{t = 1}^k\mathcal{R}_t {\phi^\mathcal{H}_t}\tps  \left ( \frac{1}{k}\sum_{t = 1}^k \phi^\mathcal{H}_t{\phi^\mathcal{H}_t}\tps  -  \frac{\gamma}{k}\sum_{t = 1}^k \phi^{\mathcal{H}}_{t+1} {\phi^\mathcal{H}_t}\tps \right ) ^{-1}
\end{align}
As the amount of data $k$ increases, the empirical covariance matrices
$\phi^{\mathcal H}_{1:k} {\phi^{\mathcal H}_{1:k}}\tps/k$ and
$\phi^{\mathcal H}_{2:k+1} {\phi^{\mathcal H}_{1:k}}\tps/k$ converge
with probability 1 to their population values, and so our estimate of
the matrix to be inverted in~(\ref{eq:LSTD}) is consistent.
Therefore, as long as this matrix is nonsingular, our estimate of the
inverse is also consistent, and our estimate of $w$ therefore
converges to the true parameters with probability 1.

\section{Predictive Features}\label{sec:pred}
Although LSTD provides a consistent estimate of the
value function parameters $w$, in practice, the potential size of the
feature vectors can be a problem. If the number of
features is large relative to the number of training
samples, then the estimation of $w$ is prone to overfitting. This
problem can be alleviated by choosing some small set of features that
only contain information that is relevant for value function
approximation. However, with the exception of LARS-TD~\cite{Kolter2009}, there
has been little work on the problem of how to select features
automatically for value function approximation when the system model is unknown; and of course, manual
feature selection depends on not-always-available expert guidance.  

We approach the problem of finding a good set of features from a \emph{bottleneck}
perspective. That is, given some signal from history, in this case a
large set of features, we would like to find a compression
that preserves only relevant information for predicting the value
function $J^\pi$. As we will see in Section~\ref{PSRs}, this
improvement is directly related to spectral identification of PSRs.

\subsection{Tests and  Features of the Future}

We first need to define precisely the task of predicting the future.
Just as a history is an ordered sequence of action-observation pairs
executed prior to time $t$, we define a \emph{test} of length $i$ to
be an ordered sequence of action-observation pairs $\tau = a_1o_1
\hdots a_io_i$ that can be executed and observed \emph{after} time
$t$~\cite{littman02}. 
The \emph{prediction} for a test $\tau$ after a history $h$, written
$\tau(h)$, is the probability that we will see the test observations
$\tau^O=o_1\ldots o_i$, given that we
\emph{intervene}~\cite{Pearl2000} to execute the test actions
$\tau^A=a_1\ldots a_i$:
\[
\tau(h) = \Pr[\tau^O\mid h, {\rm do}(\tau^A)]
\]
If $Q=\{\tau_1,\ldots,\tau_n\}$ is a set of tests, we write
$Q(h)=(\tau_1(h),\ldots,\tau_n(h))\tps$ for the corresponding vector of test
predictions.

We can generalize the notion of a test to a \emph{feature of the
  future}, a linear combination of several tests sharing a common
action sequence.  For example, if $\tau_1$ and $\tau_2$ are two tests
with $\tau_1^A=\tau_2^A\equiv\tau^A$, then we can make a feature $\phi
= 3\tau_1+\tau_2$.  This feature is \emph{executed} if we intervene to
do$(\tau^A)$, and if it is executed its \emph{value} is $3\mathbb
I(\tau_1^O)+\mathbb I(\tau_2^O)$, where $\mathbb I(o_1\ldots o_i)$
stands for an indicator random variable, taking the value 0 or 1
depending on whether we observe the sequence of observations
$o_1\ldots o_i$.  The \emph{prediction} of $\phi$ given $h$ is
$\phi(h)\equiv\mathbb E(\phi\mid
h,\mathrm{do}(\tau^A))=3\tau_1(h)+\tau_2(h)$.

While linear combinations of tests may seem restrictive, our
definition is actually very expressive: we can represent an arbitrary
function of a finite sequence of future observations.  To do so, we
take a collection of tests, each of which picks out one possible
realization of the sequence, and weight each test by the value of the
function conditioned on that realization.  For example, if our
observations are integers $1,2,\ldots,10$, we can write the square of
the next observation as $\sum_{o=1}^{10}o^2\mathbb I(o)$, and the mean
of the next two observations as
$\sum_{o=1}^{10}\sum_{o'=1}^{10}\frac{1}{2}(o+o')\mathbb I(o,o')$.

The restriction to a common action sequence is necessary: without this
restriction, all the tests making up a feature could never be executed
at once.  Once we move to feature \emph{predictions}, however, it
makes sense to lift this restriction: we will say that any linear
combination of feature predictions is also a feature prediction, even
if the features involved have different action sequences.

Action sequences raise some problems with obtaining empirical
estimates of means and covariances of features of the future: e.g., it
is not always possible to get a sample of a particular feature's value
on every time step, and the feature we choose to sample at one step
can restrict which features we can sample at subsequent
steps.  In order to carry out our derivations without running into
these problems repeatedly, we will assume for the rest of the paper
that we can reset our system after every sample, and get a new history
independently distributed as $h_t\sim\omega$ for some distribution
$\omega$.  (With some additional bookkeeping we could remove this
assumption~\cite{bowling06}, but this bookkeeping would unnecessarily
complicate our derivations.)  

Furthermore, we will introduce some new language, again to keep
derivations simple: if we have a vector of features of the future
$\phi^{\mathcal T}$, we will pretend that we can get a sample
$\phi^{\mathcal T}_t$ in which we evaluate all of our features
starting from a single history $h_t$, even if the different elements
of $\phi^{\mathcal T}$ require us to execute different action
sequences.  When our algorithms call for such a sample, we will
instead use the following trick to get a random vector with the
correct expectation (and somewhat higher variance, which doesn't
matter for any of our arguments): write $\tau^A_1,\tau_2^A,\ldots$ for
the different action sequences, and let $\zeta_1,\zeta_2,\ldots>0$ be
a probability distribution over these sequences.  We pick a single
action sequence $\tau^A_a$ according to $\zeta$, and execute
$\tau_a^A$ to get a sample $\hat\phi^{\mathcal T}$ of the features
which depend on $\tau_a^A$.  We then enter $\hat\phi^{\mathcal
  T}/\zeta_a$ into the corresponding coordinates of $\phi_t^{\mathcal
  T}$, and fill in zeros everywhere else.  It is easy to see that the
expected value of our sample vector is then correct: the probability
of selection $\zeta_a$ and the weighting factor $1/\zeta_a$ cancel
out.  We will write $\mathbb E(\phi^{\mathcal T}\mid
h_t,\text{do}(\zeta))$ to stand for this expectation.

None of the above tricks are actually necessary in our experiments
with stopping problems: we simply execute the ``continue'' action on
every step, and use only sequences of ``continue'' actions in every
test and feature.

\subsection{Finding Predictive Features Through a Bottleneck}\label{sec:predFeat}
In order to find a \emph{predictive} feature compression, we first
need to determine what we would like to predict. Since we are
interested in value function approximation, the most relevant
prediction is the value function itself; so, we could simply try to
predict total future discounted reward given a history. Unfortunately,
total discounted reward has high variance, so unless we have a lot of
data, learning will be difficult.

We can reduce variance by including other prediction tasks as well.
For example,
predicting \emph{individual} rewards at future time steps, while not
strictly necessary to predict total discounted reward, seems highly
relevant, and gives us much more immediate feedback.  Similarly,
future \emph{observations} hopefully contain information about future
reward, so trying to predict observations can help us predict
reward better.  Finally, in any specific RL application, we may be
able to add \emph{problem-specific} prediction tasks that will help
focus our attention on relevant information: for example, in a
path-planning problem, we might try to predict which of several goal
states we will reach (in addition to how much it will cost to get
there).  

We can represent all of these prediction tasks as features of the
future: e.g., to predict which goal we will reach, we add a distinct
observation at each goal state, or to predict individual rewards, we
add individual rewards as observations.\footnote{If we don't wish to
  reveal extra information by adding additional observations, we can
  instead add the corresponding feature \emph{predictions} as
  observations; these predictions, by definition, reveal no additional
  information.  To save the trouble of computing these predictions, we
  can use realized feature values rather than predictions in our
  learning algorithms below, at the cost of some extra variance: the
  expectation of the realized feature value is the same as the
  expectation of the predicted feature value.}  We will write
$\phi^{\mathcal T}_t$ for the vector of all features of the ``future
at time $t$,'' i.e., events starting at time $t+1$ and continuing
forward.


So, instead of remembering a large \emph{arbitrary} set of features of
history, we want to find a small subspace of features of history that
is relevant for predicting features of the future.  We will call this
subspace a \emph{predictive compression}, and we will write the value
function as a linear function of only the predictive compression of
features.

To find our predictive compression, we will use \emph{reduced-rank
  regression}~\cite{reinsel98}.  We define the following empirical covariance
matrices between features of the future and features of histories:
\begin{align}
 \widehat \Sigma_{\mathcal{T,H}} =\frac{1}{k} \sum_{t=1}^{k} \phi^\mathcal{T}_{t}{\phi^\mathcal{H}_{t}}\tps \qquad  \widehat \Sigma_{\mathcal{H,H}} =\frac{1}{k} \sum_{t=1}^{k} \phi^\mathcal{H}_{t}{\phi^\mathcal{H}_{t}}\tps
 \end{align}
 Let $L_\mathcal{H}$ be the lower triangular Cholesky factor of
 $\widehat \Sigma_{\mathcal{H},\mathcal{H}}$. Then we can find a
 predictive compression of histories by a singular value decomposition
 (SVD) of the weighted covariance: write
\begin{align}
\label{eq:svd}
\mathcal{U}\mathcal{D}\mathcal{V\tps} \approx  \widehat \Sigma_{\mathcal{T,H}} L_\mathcal{H}\invtps
\end{align}
for a truncated SVD~\cite{golubvanloan96book} of the weighted
covariance, where $\mathcal{U}$ are the left singular vectors,
$\mathcal{V\tps}$ are the right singular vectors, and $\mathcal{D}$ is
the diagonal matrix of singular values.  The number of columns of
$\mathcal U$, $\mathcal V$, or $\mathcal D$ is equal to the number of
retained singular values.\footnote{If our empirical
  estimate $\widehat\Sigma_{\mathcal{T,H}}$ were exact, we could keep
  all nonzero singular values to find the smallest possible compression
  that does not lose any predictive power. In practice, though, there will
  be noise in our estimate, and $\widehat \Sigma_{\mathcal{T,H}}
  L_\mathcal{H}\invtps$ will be full rank.  If we know the true rank
  $n$ of $\Sigma_{\mathcal{T,H}}$, we can choose the first $n$
  singular values to define a subspace for compression.  Or, we can
  choose a smaller subspace that results in an \emph{approximate}
  compression: by selectively dropping columns of $\mathcal U$
  corresponding to small singular values, we can trade off compression
  against predictive power.  Directions of larger variance in features
  of the future correspond to larger singular values in the SVD, so we
  minimize prediction error by truncating the smallest singular
  values.  By contrast with an SVD of the unscaled covariance, we do
  not attempt to minimize reconstruction error for features of
  history, since features of history are standardized when we multiply
  by the inverse Cholesky factor.}
Then we define
\begin{align}
\label{eq:U}
\widehat U =\mathcal{U}\mathcal{D}^{1/2}
\end{align}
to be the mapping from the low-dimensional compressed space up to the
high-dimensional space of features of the future.  


Given $\widehat U$, we would like
to find a compression operator $V$ that optimally predicts features of
the future through the bottleneck defined by $\widehat U$. The least squares estimate can be found by minimizing the loss\begin{align}
\mathcal{L}(V) = \left \| \phi_{1:k}^\mathcal{T}  - \widehat UV\phi_{1:k}^\mathcal{H}\right \|_F^2
\end{align}
where $\| \cdot \|_F$ denotes the Frobenius norm. We can find the minimum by taking the derivative of this loss with respect to $V$, setting it to zero, and solving for $V$ (see Appendix, Section~\ref{sec:Compress} for details), giving us:
\begin{align}
\widehat V &= \arg \min_{V} \mathcal{L}(V) 
=\, \widehat U\tps  \widehat \Sigma_\mathcal{T,H}( \widehat \Sigma_\mathcal{H,H})^{-1} \label{eq:V}
\end{align}

By weighting different features of the future differently, we can
change the approximate compression in interesting ways. For example,
as we will see in Section~\ref{sec:PSTDrevisited}, scaling up future
reward by a constant factor results in a \emph{value-directed
  compression}---but, unlike previous ways to find value-directed
compressions~\cite{Poupart2002}, we do not need to know a model of our
system ahead of time.  For another example, define $L_\mathcal{T}$ to
be the lower triangular Cholesky factor of the empirical covariance of
future features $\widehat \Sigma_{\mathcal{T},\mathcal{T}}$.  Then, if
we scale features of the future by $L_\mathcal{T}\invtps$, the
singular value decomposition will preserve the largest possible amount
of mutual information between features of the future and features of
history. This is equivalent to canonical correlation
analysis~\cite{Hotelling1935,Soatto01dynamicdata}, and the matrix
$\mathcal{D}$ becomes a diagonal matrix of canonical correlations
between futures and histories.

\subsection{Predictive State Temporal Difference Learning}
\label{sec:PSTD}
Now that we have found a predictive compression operator $\widehat V$
via Equation~\ref{eq:V}, we can replace the features of history
$\phi^{\mathcal H}_t$ with the compressed features $\widehat V
\phi^{\mathcal H}_t$ in the Bellman recursion,
Equation~\ref{eq:Bellman}.  Doing so results in the following
approximate Bellman equation:
\begin{align}
w\tps \widehat V\phi^\mathcal{H}_{1:k} \approx \mathcal{R}_{1:k}  + \gamma w\tps \widehat V\phi^\mathcal{H}_{2:k+1}
\end{align}
The least squares solution for $w$ is still prone to an
error-in-variables problem. The variable ${\phi^\mathcal{H}}$ is still
correlated with the true independent variables and uncorrelated with
noise, and so we can again use it as an instrumental variable to
unbias the estimate of $w$.  Define the additional empirical
covariance matrices:
\begin{align}\label{eq:RH-H+H}
\widehat \Sigma_{\mathcal{R,H}} =\frac{1}{k} \sum_{t=1}^{k} \mathcal{R}_t{\phi^\mathcal{H}_{t}}\tps \qquad \widehat \Sigma_{\mathcal{H}^+,\mathcal{H}} =\frac{1}{k} \sum_{t=1}^{k} \phi^{\mathcal{H}}_{t+1}{\phi^\mathcal{H}_{t}}\tps
\end{align}
Then, the corrected Bellman equation is:
\[\hat w\tps  \widehat V \widehat \Sigma_{\mathcal{H,H}}= \widehat \Sigma_{\mathcal{R,H}}  +  \gamma \hat w\tps \widehat V \widehat \Sigma_{\mathcal{H}^+,\mathcal{H}}
\]
and solving for $\hat w$ gives us the Predictive State Temporal Difference
(PSTD) learning algorithm:
\begin{align}
\label{eq:predictiveTD}
 \hat w\tps = \widehat \Sigma_{\mathcal{R,H}}  \left (  \widehat V \widehat \Sigma_{\mathcal{H,H}}  -  \gamma  \widehat V \widehat \Sigma_{\mathcal{H}^+,\mathcal{H}} \right ) ^\dagger
\end{align}
So far we have provided some intuition for why predictive features
should be better than arbitrary features for temporal difference
learning. Below we will show an additional benefit: the model-free
algorithm in Equation~\ref{eq:predictiveTD} is, under some
circumstances, equivalent to a model-based value function
approximation method which uses subspace identification to learn
Predictive State Representations~\cite{rosencrantz04,Boots2010b}.

\section{Predictive State Representations}
\label{PSRs} 

A predictive state representation (PSR)~\cite{littman02} is a compact
and complete description of a dynamical system. Unlike POMDPs, which
represent state as a distribution over a latent variable, PSRs
represent state as a set of predictions of tests.

Formally, a PSR consists of five elements $\langle A, O, Q, s_1, F
\rangle$. $A$ is a finite set of possible actions, and $O$ is a finite
set of possible observations.  $Q$ is a \emph{core} set of tests,
i.e., a set whose vector of predictions $Q(h)$ is a sufficient
statistic for predicting the success probabilities of \emph{all}
tests. $F$ is the set of functions $f_{\tau}$ which embody these
predictions: $\tau(h) = f_\tau(Q(h))$.  And, $m_1=Q(\epsilon)$ is the
initial prediction vector.  In this work we will restrict ourselves to
\emph{linear} PSRs, in which all prediction functions are linear:
$f_{\tau}(Q(h)) = r_\tau\tps Q(h)$ for some vector $r_{\tau} \in
\mathbb{R}^{|Q|}$. Finally, a core set $Q$ for a linear PSR is said to
be \emph{minimal} if the tests in $Q$ are linearly
independent~\cite{jaeger00a,singh04}, i.e., no one test's prediction
is a linear function of the other tests' predictions.

Since $Q(h)$ is a sufficient statistic for all tests, it is a
\emph{state} for our PSR: i.e., we can remember just $Q(h)$ instead of
$h$ itself.  After action $a$ and observation $o$, we can update
$Q(h)$ recursively: if we write $M_{ao}$ for the matrix with rows
$r_{ao\tau}\tps$ for $\tau\in Q$, then we can use Bayes' Rule to
show:%
\begin{align}
\label{update} 
Q(hao) = \frac{M_{ao} Q(h)}{\Pr[o\,|\,h, \text{do}(a)]} = \frac{M_{ao} Q(h)}{m_\infty\tps M_{ao} Q(h)}
\end{align}
where $m_\infty$ is a normalizer, defined by $m_\infty\tps Q(h)=1$ for
all $h$. 

In addition to the above PSR parameters, we need a few additional
definitions for reinforcement learning: a reward function
$\mathcal{R}(h) = \eta\tps Q(h)$ mapping predictive states to
immediate rewards, a discount factor $\gamma \in [0,1]$ which weights
the importance of future rewards vs.\ present ones, and a policy
$\pi(Q(h))$ mapping from predictive states to actions. (Specifying a
reward in terms of the core test predictions $Q(h)$ is fully general:
e.g., if we want to add a unit reward for some test $\tau\not\in Q$,
we can instead equivalently set $\eta := \eta + r_\tau$, where
$r_\tau$ is defined (as above) so that $\tau(h)=r_\tau\tps Q(h)$.)

Instead of ordinary PSRs, we will work with \emph{transformed PSRs}
(TPSRs)~\cite{rosencrantz04, Boots2010b}.  TPSRs are a generalization
of regular PSRs: a TPSR maintains a small number of sufficient
statistics which are \emph{linear combinations} of a (potentially very
large) set of test probabilities.  That is, a TPSR maintains a small
number of \emph{feature predictions} instead of test predictions.
TPSRs have exactly the same predictive abilities as regular PSRs, but
are invariant under similarity transforms: given an invertible matrix
S, we can transform $m_1\to Sm_1$, $m_\infty\tps\to m_\infty\tps
S^{-1}$, and $M_{ao}\to SM_{ao}S^{-1}$ without changing the
corresponding dynamical system, since pairs $S^{-1}S$ cancel in Eq.~\ref{update}.
The main benefit of TPSRs over regular PSRs is that, given \emph{any}
core set of tests, low dimensional parameters can be found using
spectral matrix decomposition and regression instead of combinatorial
search. In this respect, TPSRs are closely related to the transformed
representations of LDSs and HMMs found by \emph{subspace
  identification}~\cite{vanoverschee96book, katayamabook,
  Soatto01dynamicdata, zhang09}.

\subsection{Learning Transformed PSRs}
\label{sec:learning-tpsrs}

Let $Q$ be a minimal core set of tests for a dynamical system, with
cardinality $n = |Q|$ equal to the linear dimension of the
system. Then, let $\mathcal{T}$ be a larger core set of tests (not
necessarily minimal, and possibly even with $|\mathcal T|$ countably
infinite).  And, let $\mathcal{H}$ be the set of all possible
histories. ($|\mathcal{H}|$ is finite or countably infinite, depending
on whether our system is finite-horizon or infinite-horizon.)





As before, write $\phi^{\mathcal H}_t\in\Re^\ell$ for a vector of
features of history at time $t$, and write $\phi^{\mathcal
  T}_t\in\Re^\ell$ for a vector of features of the future at time $t$.
Since $\mathcal T$ is a core set of tests, by definition we can
compute any test prediction $\tau(h)$ as a linear function of
$\mathcal T(h)$.  And, since feature predictions are linear
combinations of test predictions, we can also compute any feature
prediction $\phi(h)$ as a linear function of $\mathcal T(h)$.
We define the matrix $\Phi^\mathcal{T}\in\Re^{\ell\times|\mathcal{T}|}$ to embody
our predictions of future features: that is, an entry of
$\Phi^\mathcal{T}$ is the weight of one of the tests in $\mathcal T$
for calculating the {prediction} of one of the features in
$\phi^{\mathcal T}$.


Below we define several covariance matrices,
Equation~\ref{eq:PSRcovs}(a--d), in terms of the observable quantities
$\phi^\mathcal{T}_t$, $\phi^\mathcal{H}_t$, $a_t$, and $o_t$, and show
how these matrices relate to the parameters of the underlying PSR.
These relationships then lead to our learning algorithm,
Eq.~\ref{eq:psrParams} below.

First we define $\Sigma_{\mathcal{H},\mathcal{H}} $, the covariance matrix of features of histories, as $ \mathbb E[\phi^\mathcal{H}_t{\phi^\mathcal{H}_t}\tps\mid h_t \sim \omega]$. Given $k$ samples, we can approximate this covariance:
\begin{subequations}\label{eq:PSRcovs}
\begin{align}
[\widehat \Sigma_{\mathcal{H},\mathcal{H}}]_{i,j} = \frac{1}{k}\sum_{t = 1}^{k} \phi^\mathcal{H}_{it}\phi^\mathcal{H}_{jt} \implies \widehat \Sigma_{\mathcal{H},\mathcal{H}} = \frac{1}{k}\phi^\mathcal{H}_{1:k} {\phi^\mathcal{H}_{1:k}}\tps.
\end{align}
\mbox{As $k \rightarrow \infty$, the empirical covariance $\widehat \Sigma_{\mathcal{H},\mathcal{H}}$ converges to the true covariance $\Sigma_{\mathcal{H},\mathcal{H}}$ with  probability 1.}

Next we define $\Sigma_{\mathcal S,\mathcal H}$, the cross covariance
of states and features of histories.  Writing $s_t=Q(h_t)$ for the
(unobserved) state at time $t$, let
\[
\Sigma_{\mathcal S,\mathcal H} = \mathbb E\left[
\frac{1}{k}s_{1:k}{\phi^{\mathcal H}_{1:k}}\tps
\,\middle |\, h_t\sim\omega\,(\forall t)
\right]
\]
We cannot directly estimate $\Sigma_{\mathcal S,\mathcal H}$ from
data, but this matrix will appear as a factor in several of the
matrices that we define below.

Next we define $\Sigma_{\mathcal{T},\mathcal{H}}$, the cross
covariance matrix of the features of tests and histories:
$\Sigma_{\mathcal{T},\mathcal{H}} \equiv \mathbb
E[\phi^\mathcal{T}_{t}{\phi^\mathcal{H}_{t}}\tps \mid h_t \sim
\omega,\text{do}(\zeta)]$. The true covariance is the expectation of
the sample covariance $\widehat \Sigma_\mathcal{T,H}$:
\begin{align}
[\widehat \Sigma_\mathcal{T,H}]_{i,j} &\equiv \frac{1}{k}\sum_{t = 1}^k \phi^\mathcal{T}_{i,t}\phi_{j,t}^\mathcal{H}  \nonumber\\
[\Sigma_\mathcal{T,H}]_{i,j} &= \mathbb{E}\left [ \frac{1}{k}\sum_{t = 1}^k \phi_{i,t}^\mathcal{T}\phi_{j,t}^\mathcal{H}  \,\middle |\,  h_t \sim \omega\, (\forall t) ,\text{do}(\zeta)\,(\forall t)\right]\nonumber\\
&= \mathbb{E}\left [ \frac{1}{k}\sum_{t = 1}^k \mathbb{E}\left[ \phi_{i,t}^\mathcal{T}\mid h_t ,\text{do}(\zeta)\right]\phi_{j,t}^\mathcal{H}  \,\middle |\,  h_t \sim \omega\, (\forall t) ,\text{do}(\zeta)\,(\forall t)\right]\nonumber\\
%
%
&=\mathbb{E}\left [  \frac{1}{k}\sum_{t=1}^k \sum_{\tau \in \mathcal{T}}\Phi^\mathcal{T}_{i,\tau}\tau(h_t)\phi^\mathcal{H}_{j,t}  \middle |\, h_t\sim \omega \, (\forall t)\right]\nonumber\\
&=\mathbb{E}\left [  \frac{1}{k}\sum_{t=1}^k \sum_{\tau \in
    \mathcal{T}} \Phi^\mathcal{T}_{i,\tau}r_\tau\tps Q(h_t)\phi^\mathcal{H}_{j,t}  \middle |\, h_t\sim \omega \, (\forall t) \right]\nonumber\\
&=\sum_{\tau \in \mathcal{T}}\Phi^\mathcal{T}_{i,\tau}r_\tau\tps\, \mathbb{E}\left [ \frac{1}{k}\sum_{t=1}^kQ(h_t)\phi^\mathcal{H}_{j,t} \middle |\, h_t\sim \omega \, (\forall t) \right] \nonumber\\
&=\sum_{\tau \in \mathcal{T}}\Phi^\mathcal{T}_{i,\tau}r_\tau\tps\, \mathbb{E}\left [ \frac{1}{k}\sum_{t=1}^k s_t\phi^\mathcal{H}_{j,t} \middle |\, h_t\sim \omega \, (\forall t) \right] \nonumber\\
\implies \Sigma_{\mathcal{T},\mathcal{H}} &=  \Phi^\mathcal{T} R
\Sigma_{\mathcal S, \mathcal H}\label{SigmaTH}
\end{align}  
where the vector $r_{\tau}$ is the linear function that specifies the
probability of the test $\tau$ given the probabilities of tests in the
core set $Q$, and the matrix $R$ has all of the $r_\tau$
vectors as rows.

The above derivation shows that, because of our assumptions about the
linear dimension of the system, the matrix
$\Sigma_{\mathcal{T},\mathcal{H}}$ has factors $R \in
\mathbb{R}^{|\mathcal T|\times n}$ and $\Sigma_{\mathcal S, \mathcal H} \in
\mathbb{R}^{n\times\ell}$. Therefore, the \emph{rank} of
$\Sigma_{\mathcal{T},\mathcal{H}}$ is no more than $n$, the linear
dimension of the system.  
We can also see that,
since the size of $\Sigma_{\mathcal T, \mathcal H}$ is fixed but the
number of samples $k$ is increasing,
the empirical covariance $\widehat
\Sigma_{\mathcal{T},\mathcal{H}}$ converges to the true covariance
$\Sigma_{\mathcal{T},\mathcal{H}}$ with probability 1.

Next we define $\Sigma_{\mathcal{H},ao,\mathcal{H}}$,
a \emph{set} of matrices, one for each action-observation pair, that
represent the covariance between features of history before and after
taking action $a$ and observing $o$.  In the following, $\mathbb
I_t(o)$ is an indicator variable for whether we see observation $o$ at
step $t$.
\begin{align}
\widehat \Sigma_{\mathcal{H},ao,\mathcal{H}} &\equiv
\frac{1}{k}\sum_{t = 1}^k \phi_{t+1}^\mathcal{H}\mathbb
I_t(o){\phi_{t}^\mathcal{H}}\tps  \nonumber\\ 
\Sigma_{\mathcal{H},ao,\mathcal{H}} &\equiv
\mathbb{E}\left[\widehat \Sigma_{\mathcal{H},ao,\mathcal{H}}
  \, \middle | \, h_t \sim \omega \, (\forall t),\,
  \text{do}(a)\,(\forall t)\right] \nonumber\\
&= \mathbb{E}\left [ \frac{1}{k}\sum_{t = 1}^k
  \phi_{t+1}^\mathcal{H}\mathbb I_t(o){\phi_{t}^\mathcal{H}}\tps \, \middle | \, h_t \sim \omega \, (\forall t),\, \text{do}(a)\,(\forall t)\right]
\end{align}
Since the dimensions of each $\widehat
\Sigma_{\mathcal{H},ao,\mathcal{H}}$ are fixed, as $k \rightarrow
\infty$ these empirical covariances converge to the true covariances
$\Sigma_{\mathcal{H},ao,\mathcal{H}}$ with probability 1.

Finally we define $\Sigma_\mathcal{R,H} \equiv \mathbb
E[\mathcal{R}_t{\phi^{\mathcal H}_t}\tps\mid h_t \sim \omega] $, and approximate the covariance (in this case a vector) of reward and features of history:
\begin{align}
\widehat \Sigma_\mathcal{R,H} &\equiv \frac{1}{k}\sum_{t = 1}^k
\mathcal R_t{\phi_{t}^\mathcal{H}}\tps  \nonumber\\
\Sigma_\mathcal{R,H} &\equiv \mathbb{E}\left[\widehat \Sigma_\mathcal{R,H} \,\middle |\,  h_t \sim \omega\, (\forall t)\right]\nonumber\\
&= \mathbb{E}\left [ \frac{1}{k}\sum_{t = 1}^k \mathcal R_t{\phi_{t}^\mathcal{H}}\tps  \,\middle |\,  h_t \sim \omega\, (\forall t)\right]\nonumber\\
&=\mathbb{E}\left [  \frac{1}{k}\sum_{t=1}^k \eta\tps Q(h_t){\phi^\mathcal{H}_{t}}\tps\, \middle |\, h_t\sim \omega \, (\forall t)\right]\nonumber\\
&=\eta\tps\mathbb{E}\left [  \frac{1}{k}\sum_{t=1}^k s_t{\phi^\mathcal{H}_{t}}\tps \,\middle |\, h_t\sim \omega \, (\forall t)\right]\nonumber\\
%
&= \eta\tps\Sigma_{\mathcal S,\mathcal H}
\end{align}
\end{subequations}
Again, as $k \rightarrow \infty$, $\widehat
\Sigma_\mathcal{R,H}$  converges to $\Sigma_\mathcal{R,H}$  with
probability 1.

We now wish to use the above-defined matrices to learn a TPSR from
data.  To do so we need to make a somewhat-restrictive assumption: we
assume that our features of history are rich enough to determine the
state of the system, i.e., the regression from $\phi^{\mathcal H}$ to
$s$ is exact: $s_t = \Sigma_{\mathcal S,\mathcal H}\Sigma_{\mathcal
  H,\mathcal H}^{-1}\phi_t^{\mathcal H}$.  We discuss how to relax
this assumption below in Section~\ref{sec:insights}.  We also need a
matrix $U$ such that $U\tps \Phi^\mathcal{T} R$ is invertible; with
probability 1 a random matrix satisfies this condition, but as we will
see below, it is useful to choose $U$ via SVD of a scaled version of
$\Sigma_{\mathcal T, \mathcal H}$ as described in
Sec.~\ref{sec:predFeat}.

Using our assumptions we can show a useful identity for
$\Sigma_{\mathcal H,ao,\mathcal H}$:
\begin{align}
\Sigma_{\mathcal S,\mathcal H}\Sigma_{\mathcal
  H,\mathcal H}^{-1}\Sigma_{\mathcal H,ao,\mathcal H} 
&=
 \mathbb{E}\left [ \frac{1}{k}\sum_{t = 1}^k
\Sigma_{\mathcal S,\mathcal H}\Sigma_{\mathcal
  H,\mathcal H}^{-1}  \phi_{t+1}^\mathcal{H}\mathbb
I_t(o){\phi_{t}^\mathcal{H}}\tps \,
  \middle | \, h_t \sim \omega \, (\forall t),\,
  \text{do}(a)\,(\forall t)\right]\nonumber\\
&=
\mathbb{E}\left [ \frac{1}{k}\sum_{t = 1}^k
  s_{t+1}\mathbb I_t(o){\phi_{t}^\mathcal{H}}\tps \,
  \middle | \, h_t \sim \omega \, (\forall t),\,
  \text{do}(a)\,(\forall t)\right]\nonumber\\
&=
\mathbb{E}\left [ \frac{1}{k}\sum_{t = 1}^k
  M_{ao}s_{t}{\phi_{t}^\mathcal{H}}\tps \,
  \middle | \, h_t \sim \omega \, (\forall t)\right]\nonumber\\
&=
\label{eq:HaoH-Mao}
M_{ao}\Sigma_{\mathcal S,\mathcal H}
\end{align}
This identity is at the heart of our learning algorithm: it shows that
$\Sigma_{\mathcal H,ao,\mathcal H}$ contains a hidden copy of
$M_{ao}$, the main TPSR parameter that we need to learn.  We would
like to recover $M_{ao}$ via Eq.~\ref{eq:HaoH-Mao},
$M_{ao}=\Sigma_{\mathcal S,\mathcal H}\Sigma_{\mathcal H,\mathcal
  H}^{-1}\Sigma_{\mathcal H,ao,\mathcal H}\Sigma_{\mathcal S,\mathcal
  H}^\dag$; but of course we do not know $\Sigma_{\mathcal S,\mathcal
  H}$.  Fortunately, though, it turns out that we can use $U\tps
\Sigma_{\mathcal T,\mathcal H}$ as a stand-in, as described below,
since this matrix differs from $\Sigma_{\mathcal S,\mathcal H}$ only by an
invertible transform (Eq.~\ref{SigmaTH}).

We now show how to recover a TPSR from the matrices
$\Sigma_{\mathcal{T},\mathcal{H}}$,
$\Sigma_{\mathcal{H},\mathcal{H}}$,
$\Sigma_{\mathcal{R},\mathcal{H}}$,
$\Sigma_{\mathcal{H},ao,\mathcal{H}}$, and $U$.  Since a TPSR's
predictions are invariant to a similarity transform of its parameters,
our algorithm only recovers the TPSR parameters to within a similarity
transform.
\begin{subequations}\label{eq:psrParams}
\begin{align}
b_t &\equiv U\tps \Sigma_\mathcal{T,H}(\Sigma_\mathcal{H,H})^{-1}\phi^\mathcal{H}_t \nonumber\\
&= U\tps\Phi^\mathcal{T} R \Sigma_{\mathcal S, \mathcal H} (\Sigma_\mathcal{H,H})^{-1}\phi^\mathcal{H}_t\nonumber\\ 
%
&= (U\tps \Phi^{\mathcal{T}}R) s_t\\
B_{ao} &\equiv U\tps \Sigma_\mathcal{T,H}(\Sigma_\mathcal{H,H})^{-1}\Sigma_{\mathcal{H},ao,\mathcal{H}}(U\tps \Sigma_{\mathcal{T},\mathcal{H}})^\dag \nonumber\\
&= U\tps \Phi^\mathcal{T} R \Sigma_{\mathcal S,\mathcal H}(\Sigma_\mathcal{H,H})^{-1}\Sigma_{\mathcal{H},ao,\mathcal{H}}(U\tps \Sigma_{\mathcal{T},\mathcal{H}})^\dag \nonumber\\
&= (U\tps\Phi^\mathcal{T} R) M_{ao}\, \Sigma_{\mathcal S, \mathcal H}(U\tps \Sigma_{\mathcal{T},\mathcal{H}})^\dag\nonumber\\
& = (U\tps\Phi^\mathcal{T} R) M_{ao}(U\tps\Phi^{\mathcal{T}}R)^{-1}(U\tps\Phi^{\mathcal{T}}R)\Sigma_{\mathcal S, \mathcal H}(U\tps \Sigma_{\mathcal{T},\mathcal{H}})^\dag\nonumber\\
&= (U\tps \Phi^\mathcal{T} R) M_{ao}(U\tps\Phi^{\mathcal{T}}R)^{-1}\\
b_\eta \tps &\equiv \Sigma_\mathcal{R,H} (U\tps \Sigma_\mathcal{T,H})^\dag\nonumber\\
&= \eta\tps\Sigma_{\mathcal S, \mathcal H}(U\tps \Sigma_\mathcal{T,H})^\dag\nonumber\\
&=  \eta\tps(U\tps \Phi^{\mathcal{T}}R)^{-1}(U\tps \Phi^{\mathcal{T}}R)\Sigma_{\mathcal S, \mathcal H}(U\tps \Sigma_\mathcal{T,H})^\dag\nonumber\\
&= \eta\tps(U\tps\Phi^{\mathcal{T}}R)^{-1}
\end{align}
\end{subequations}
Our PSR learning algorithm is simple: simply replace each true
covariance matrix in Eq.~\ref{eq:psrParams} by its empirical estimate.
Since the empirical estimates converge to their true values with
probability 1 as the sample size increases, our learning algorithm is
clearly statistically consistent.

\subsection{Predictive State Temporal Difference Learning (Revisited)}\label{sec:PSTDrevisited}


Finally, we are ready to show that the model-free PSTD learning
algorithm introduced in Section~\ref{sec:PSTD} is \emph{equivalent} to
a model-based algorithm built around PSR learning.
For a fixed policy $\pi$, a TPSR's value function is a linear function
of state, $J^\pi(s)=w\tps b$, and is the solution of the TPSR Bellman
equation~\cite{james06}: for all $b$, $w\tps b = b_\eta\tps b +
\gamma\sum_{o\in O} w\tps B_{\pi o} b$, or equivalently,
\[
w\tps = b_\eta\tps + \gamma \sum_{o \in O} w\tps B_{\pi o}
\]
If we substitute in our learned PSR parameters
from Equations~\ref{eq:psrParams}(a--c), we get
\begin{align}
\hat w\tps &= \widehat\Sigma_\mathcal{R,H} (U\tps \widehat\Sigma_\mathcal{T,H})^\dag +
\gamma \sum_{o \in O}\hat w\tps U\tps
\widehat\Sigma_\mathcal{T,H}(\widehat\Sigma_\mathcal{H,H})^{-1}\widehat\Sigma_{\mathcal{H},\pi
  o,\mathcal{H}}(U\tps
\widehat\Sigma_{\mathcal{T},\mathcal{H}})^\dag\nonumber\\
\hat w\tps U\tps \widehat\Sigma_\mathcal{T,H} &= \widehat\Sigma_\mathcal{R,H}  +
\gamma \hat w\tps U\tps
\widehat\Sigma_\mathcal{T,H}(\widehat\Sigma_\mathcal{H,H})^{-1}\widehat\Sigma_{\mathcal{H}^+,\mathcal{H}}\nonumber
\end{align}
since, by comparing Eqs.~\ref{eq:PSRcovs}c and~\ref{eq:RH-H+H}, we can
see that $\sum_{o \in O}\widehat\Sigma_{\mathcal{H},\pi o,\mathcal{H}}=
\widehat\Sigma_{\mathcal{H}^+,\mathcal{H}}$.  Now, suppose that we define
$\widehat U$ and $\widehat V$ by Eqs.~\ref{eq:U} and~\ref{eq:V}, and
let $U=\widehat U$ as suggested above in
Sec.~\ref{sec:learning-tpsrs}.  Then $U\tps\widehat\Sigma_\mathcal{T,H} =
\widehat V\widehat\Sigma_\mathcal{H,H}$, and
\begin{align}
\hat w\tps\widehat V\widehat\Sigma_\mathcal{H,H}
 &= \widehat\Sigma_\mathcal{R,H} +
\gamma \hat w\tps\widehat V\widehat\Sigma_{\mathcal{H}^+,\mathcal{H}}\nonumber\\
\hat w\tps
 &= \widehat\Sigma_\mathcal{R,H}\left(\widehat V\widehat\Sigma_\mathcal{H,H}-
\gamma \widehat V\widehat\Sigma_{\mathcal{H}^+,\mathcal{H}}\right)^\dag\label{eq:pstd2}
\end{align}
Eq.~\ref{eq:pstd2} is exactly the PSTD algorithm
(Eq.~\ref{eq:predictiveTD}).  So, we have shown that, if we learn a
PSR by the subspace identification algorithm of
Sec.~\ref{sec:learning-tpsrs} and then compute its value function via
the Bellman equation, we get the exact same answer as if we had
directly learned the value function via the
model-free PSTD method.  In addition to adding to our
understanding of both methods, an important corollary of this result
is that PSTD is a \emph{statistically consistent} algorithm
for PSR value function approximation---to our knowledge, the first
such result for a TD method.

PSTD learning is related to value-directed compression of
POMDPs~\cite{Poupart2002}.  If we learn a TPSR from data generated by
a POMDP, then the TPSR state is exactly a linear compression of the
POMDP state~\cite{singh04,rosencrantz04}.  The compression can be
exact or approximate, depending on whether we include enough features
of the future and whether we keep all or only some nonzero singular
values in our bottleneck.  If we include \emph{only} reward as
a feature of the future, we get a value-directed compression in the
sense of Poupart and Boutilier~\cite{Poupart2002}.  If desired, we can
\emph{tune} the degree of value-directedness of our compression by
scaling the relative variance of our features: the higher the variance
of the reward feature compared to other features, the more
value-directed the resulting compression will be.  Our work
significantly diverges from previous work on POMDP compression in one
important respect: prior work assumes access to the true POMDP model,
while we make no such assumption, and learn a compressed
representation directly from data.

\subsection{Insights from Subspace Identification}\label{sec:insights}
The close connection to subspace identification for PSRs provides
additional insight into the temporal difference learning procedure. In
Equation~\ref{eq:psrParams} we made the assumption that the features
of history are rich enough to completely determine the state of the
dynamical system. In fact, using theory developed
in~\cite{Boots2010b}, it is possible to relax this assumption and
instead assume that state is merely \emph{correlated} with features of
history. In this case, we need to introduce a new set of covariance
matrices $\Sigma_{\mathcal{T},ao,\mathcal{H}} \equiv
\mathbb{E}[\phi^\mathcal{T}_t \mathbb I_t(o)
{\phi^\mathcal{H}_t}\tps\mid h_t\sim \omega, \text{do}(a,\zeta)]$, one for each
action-observation pair, that represent the covariance between
features of history before and features of tests after taking action
$a$ and observing $o$. We can then estimate the TPSR transition
matrices as $\widehat B_{ao} = \widehat U\tps \widehat
\Sigma_{\mathcal{T},ao,\mathcal{H}}(\widehat U \tps \widehat
\Sigma_\mathcal{T,H} )^\dag$ (see~\cite{Boots2010b} for proof
details). The value function parameter $w$ can be estimated as $\hat
w\tps = \widehat \Sigma_\mathcal{R,H}(\widehat U \tps \widehat
\Sigma_\mathcal{T,H} )^\dag(I - \sum_{o\in O}\widehat U\tps \widehat
\Sigma_{\mathcal{T},ao,\mathcal{H}}(\widehat U \tps \widehat
\Sigma_\mathcal{T,H} )^\dag)^\dag = \widehat
\Sigma_\mathcal{R,H}(\widehat U \tps \widehat \Sigma_\mathcal{T,H} -
\sum_{o\in O}\widehat U\tps \widehat
\Sigma_{\mathcal{T},ao,\mathcal{H}})^\dag$ (the proof is similar to
Equation~\ref{eq:pstd2}). Since we no longer assume that state is
completely specified by features of history, we can no longer apply the learned
value function to $\widehat U
\Sigma_\mathcal{T,H}(\Sigma_\mathcal{H,H})^{-1}\phi_t$ at each time
$t$. Instead we need to learn a full PSR model and \emph{filter} with
the model to estimate state. Details on this procedure can be found
in~\cite{Boots2010b}.

\section{Experimental Results}
We designed several experiments to evaluate the properties of the PSTD
learning algorithm. In the first set of experiments we look at the
comparative merits of PSTD with respect to LSTD and LARS-TD when
applied to the problem of estimating the value function of a
reduced-rank POMDP\@. In the second set of experiments, we apply PSTD
to a benchmark optimal stopping problem (pricing a
fictitious financial derivative), and show that PSTD outperforms
competing approaches.

\subsection{Estimating the Value Function of a RR-POMDP}
\label{sec:RR-POMDP}
We evaluate the PSTD learning algorithm on a synthetic example derived
from~\cite{Siddiqi10a}. The problem is to find the value function of a
policy in a partially observable Markov decision Process (POMDP). The
POMDP has 4 latent states, but the policy's transition matrix is low
rank: the resulting belief distributions can be represented in a
3-dimensional subspace of the original belief simplex. A reward of $1$
is given in the first and third latent state and a reward of $0$ in
the other two latent states (see Appendix,
Section~\ref{sec:AppendExp}). The system emits 2 possible
observations, conflating information about the latent states.

We perform 3 experiments, comparing the performance of LSTD, LARS-TD, PSTD, and PSTD as formulated in Section~\ref{sec:insights} (which we call PSTD2) when different sets of features are used. In each case we compare the value function estimated by each algorithm to the true value function computed by $J^\pi = \mathcal{R}(I -\gamma T^\pi)^{-1}$.

\begin{figure*}[!tb]
\includegraphics[width=1.0\linewidth]{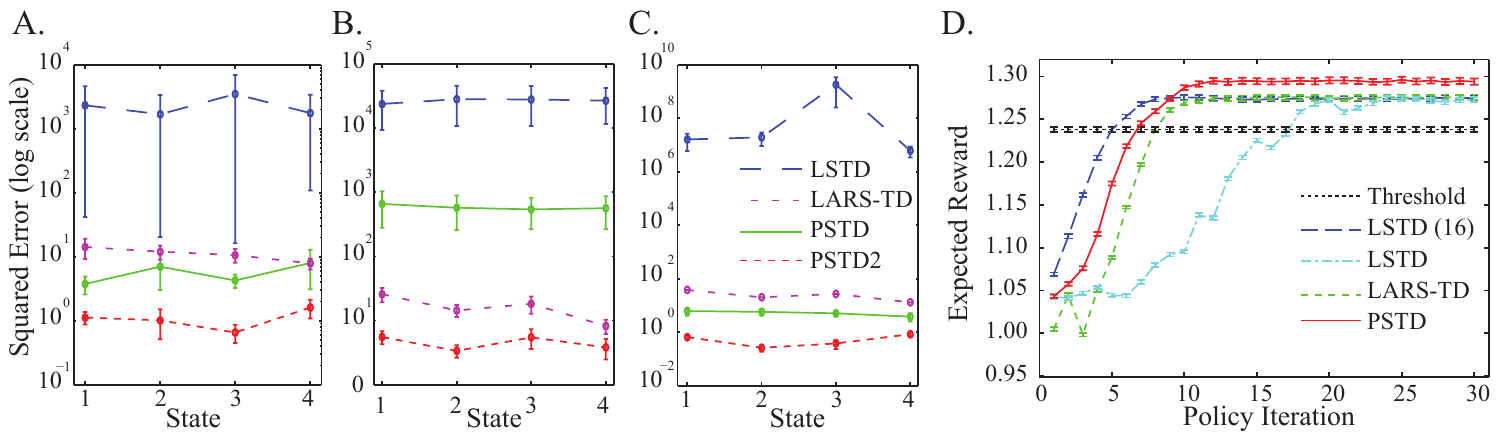}
\caption{Experimental Results. Error bars indicate standard error. (A)
  Estimating the value function with a small number of informative
  features. PSTD and PSTD2 both do well. (B) Estimating the value
  function with a small set of informative features and a large set of
  random features. LARS-TD is designed for this scenario and
  dramatically outperforms PSTD and LSTD, however it does not outperform PSTD2. (C) Estimating the value
  function with a large set of semi-informative features. PSTD is able
  to determine a small set of compressed features that retain the
  maximal amount of information about the value function,
  outperforming LSTD by a very large margin. (D) Pricing a high-dimensional
  derivative via policy iteration. The y-axis is expected reward for
  the current policy at each iteration. The optimal threshold strategy
  (sell if price is above a threshold~\cite{Tsitsiklis97}) is in
  black, LSTD (16 canonical features) is in blue,  LSTD (on the full
  220 features) is cyan, LARS-TD (feature selection from set of 220)
  is in green, and PSTD (16 dimensions, compressing 220 features (16 +
  204)) is in red.}\label{fig:results}
\end{figure*}

In the first experiment we execute the policy $\pi$ for 1000 time
steps. We split the data into overlapping histories
and tests of length 5, and sample 10 of these histories and tests to
serve as centers for Gaussian radial basis functions.  We then
evaluate each basis function at every remaining sample. Then, using
these features, we learned the value function using LSTD, LARS-TD,
PSTD with linear dimension 3, and PSTD2 with linear dimension 3
(Figure~\ref{fig:results}(A)).\footnote{Comparing LSTD and PSTD is
  straightforward; the two methods differ only by the compression
  operator $\widehat V$.} In this experiment, PSTD and PSTD2 both had lower mean squared error than the other approaches.  For the second experiment, we added 490 random features to
the 10 good features and then attempted to learn the value function
with each of the 3 algorithms (Figure~\ref{fig:results}(B)). In this
case, LSTD and PSTD both had difficulty fitting the value function due
to the large number of irrelevant features in both tests and histories
and the relatively small amount of training data. LARS-TD, designed
for precisely this scenario, was able to select the 10 relevant
features and estimate the value function better by a substantial
margin. Surprisingly, in this experiment PSTD2 not only outperformed PSTD but bested even LARS-TD.  For the third experiment, we increased the number of sampled
features from 10 to 500. In this case, each feature was somewhat
relevant, but the number of features was relatively large compared to
the amount of training data. This situation occurs frequently in
practice: it is often easy to find a large number of features that are
at least somewhat related to state. PSTD and PSTD2 both outperform LARS-TD and each of these subspace and subset selection methods outperform LSTD by a large margin by \emph{efficiently} estimating the
value function (Figure~\ref{fig:results}(C)).

\subsection{Pricing A High-dimensional Financial Derivative}
Derivatives are financial contracts with payoffs linked to the future
prices of basic assets such as stocks, bonds and commodities.  In some
derivatives the contract holder has no choices, but in more complex
cases, the contract owner must make decisions---e.g., with \emph{early
  exercise} the contract holder can decide to terminate the contract
at any time and receive payments based on prevailing market
conditions.  In these cases, the value of the derivative depends on
how the contract holder acts. Deciding when to exercise is therefore
an optimal stopping problem: at each point in time, the contract
holder must decide whether to continue holding the contract or
exercise.  Such stopping problems provide an ideal testbed for policy
evaluation methods, since we can easily collect a single data set
which is sufficient to evaluate any policy: we just choose the
``continue'' action forever.  (We can then evaluate the ``stop'' action
easily in any of the resulting states, since the immediate reward is
given by the rules of the contract, and the next state is the terminal
state by definition.)

We consider the financial derivative introduced by Tsitsiklis and Van
Roy~\cite{Tsitsiklis97}. The derivative generates payoffs that are
contingent on the prices of a single stock. At the end of a given day,
the holder may opt to exercise. At exercise the owner receives a
payoff equal to the current price of the stock divided by the price
100 days beforehand.  We can think of this derivative as a ``psychic
call'': the owner gets to decide whether s/he would like to have bought
an ordinary 100-day European call option, at the then-current market
price, 100 days ago.  

In our simulation (and unknown to the investor), the underlying stock
price follows a geometric Brownian motion with volatility $\sigma =
0.02$ and continuously compounded short term growth rate $\rho =
0.0004$.  Assuming stock prices fluctuate only on days when the market
is open, these parameters correspond to an annual growth rate of $\sim
10\%$.  In more detail, if $w_t$ is a standard Brownian motion, then
the stock price $p_t$ evolves as $\nabla p_t = \rho p_t \nabla t +
\sigma p_t \nabla w_t$, and we can summarize relevant state at the end
of each day as a vector $x_t \in
\mathbb{R}^{100}$, with $x_t = \left( \frac{p_{t-99}}{p_{t-100}},
  \frac{p_{t-98}}{p_{t-100}},\hdots,\frac{p_{t}}{p_{t-100}}
\right)\tps$.  The $i$th dimension $x_t(i)$ represents the amount a
\$1 investment in a stock at time $t-100$ would grow to at time $t-100
+ i$. This process is Markov and ergodic~\cite{Tsitsiklis97,Choi2006}:
$x_t$ and $x_{t+100}$ are independent and identically distributed. The
immediate reward for exercising the option is $G(x) = x(100)$, and the
immediate reward for continuing to hold the option is 0.  The discount
factor $\gamma = e^{-\rho}$ is determined by the growth rate; this
corresponds to assuming that the risk-free interest rate is
equal to the stock's growth rate, meaning that the investor gains
nothing in expectation by holding the stock itself. 

The value of the derivative, if the current state is $x$, is given by
$V^*(x)=\sup_{t}\mathbb{E}[\gamma^t G(x_t)\mid x_0=x]$. Our goal is to
calculate an approximate value function $V(x)=w\tps\phi^{\mathcal
  H}(x)$, and then use this value
function to generate a stopping time $\min\{t \, | \, G(x_t) \geq
V(x_t)\}$. To do so, we sample a sequence of 1,000,000 states $x_t \in
\mathbb{R}^{100}$ and calculate features $\phi^{\mathcal H}$ of each
state.  We then perform \emph{policy iteration} on this sample,
alternately estimating the value function under a given policy and
then using this value function to define a new greedy policy ``stop if
$G(x_t) \geq w\tps \phi^{\mathcal{H}}(x_t)$.''

Within the above strategy, we have two main choices: which features do
we use, and how do we estimate the value function in terms of these
features.  For value function estimation, we used LSTD, LARS-TD, or
PSTD\@.  In each case we re-used our 1,000,000-state sample trajectory
for all iterations: we start at the beginning and follow the
trajectory as long as the policy chooses the ``continue'' action, with
reward 0 at each step.  When the policy executes the ``stop'' action,
the reward is $G(x)$ and the next state's features are all 0; we then
restart the policy 100 steps in the future, after the process has
fully mixed. For feature selection, we are fortunate: previous researchers have hand-selected a ``good'' set of 16 features 
for this data set through repeated trial and error (see Appendix,
Section~\ref{sec:AppendExp} and~\cite{Tsitsiklis97,Choi2006}). We
greatly expand this set of features, then use PSTD to synthesize a
small set of high-quality combined features. Specifically, we add the entire 100-step
state vector, the squares of the components of the state vector, and several
additional nonlinear features, increasing the total number of features
from 16 to 220. We use histories of length 1, tests
of length 5, and (for comparison's sake) we choose a linear dimension
of 16. Tests (but not histories) were value-directed by reducing the variance of all
features \emph{except reward} by a factor of 100.

Figure~\ref{fig:results}D shows results.  We compared PSTD (reducing
220 to 16 features) to LSTD with either the 16 hand-selected features
or the full 220 features, as well as to LARS-TD (220 features) and to
a simple thresholding strategy~\cite{Tsitsiklis97}.  In each case we
evaluated the final policy on 10,000 new random trajectories.
PSTD outperformed each of its competitors, improving on the next best
approach, LARS-TD, by 1.75 percentage points. In fact, PSTD performs
better than the best previously reported
approach~\cite{Tsitsiklis97,Choi2006} by 1.24 percentage points.
These improvements correspond to appreciable fractions of the
risk-free interest rate (which is about 4 percentage points over the
100 day window of the contract), and therefore to significant
arbitrage opportunities: an investor who doesn't know the best
strategy will consistently undervalue the security, allowing an
informed investor to buy it for below its expected value.




\section{Conclusion}

In this paper, we attack the feature selection problem for temporal
difference learning.  Although well-known temporal difference
algorithms such as LSTD can provide asymptotically unbiased estimates
of value function parameters in linear architectures, they can have
trouble in finite samples: if the number of features is large relative
to the number of training samples, then they can have high variance in
their value function estimates.  For this reason, in real-world
problems, a substantial amount of time is spent selecting a small set
of features, often by trial and error~\cite{Tsitsiklis97,Choi2006}.

To remedy this problem, we present the PSTD algorithm, a new approach
to feature selection for TD methods, which demonstrates how insights
from system identification can benefit reinforcement learning.  PSTD
automatically chooses a small set of features that are relevant for
prediction and value function approximation.  It approaches feature
selection from a \emph{bottleneck} perspective, by finding a small set
of features that preserves only \emph{predictive} information.
Because of the focus on predictive information, the PSTD approach is
closely connected to PSRs: under appropriate assumptions, PSTD's
compressed set of features is asymptotically equivalent to TPSR state,
and PSTD is a consistent estimator of the PSR value function.

We demonstrate the merits of PSTD compared to two popular
alternative algorithms, LARS-TD and LSTD, on a synthetic example, and
argue that PSTD is most effective when approximating a value function
from a large number of features, each of which contains at least a
little information about state. Finally, we apply PSTD to a
difficult optimal stopping problem, and demonstrate the practical
utility of the algorithm by outperforming several alternative
approaches and topping the best reported previous results.

\section*{Acknowledgements}
Byron Boots was supported by the NSF under grant number EEEC-0540865. Byron Boots and Geoffrey J. Gordon were supported by ONR MURI grant number N00014-09-1-1052.

\bibliographystyle{unsrt}
\bibliography{PSTDarXiv}

\clearpage
\appendix
 
\section*{Appendix}

\section{Determining the Compression Operator}
\label{sec:Compress}
We find a compression operator $V$ that optimally predicts test-features through the CCA bottleneck defined by $\widehat U$. The least squares estimate can be found by minimizing the following loss\begin{align*}
\mathcal{L}(V) &= \left \| \phi_{1:k}^\mathcal{T}  - \widehat UV\phi_{1:k}^\mathcal{H}\right \|_F^2 \\
\widehat V &= \arg \min_{V} \mathcal{L}(V) 
\end{align*}
where $\| \cdot \|_F$ denotes the Frobenius norm. We can find $\widehat V$ by taking a derivative of this loss $\mathcal{L}$ with respect to $V$, setting it to zero, and solving for $V$
\begin{align*}
\mathcal{L} &= \frac{1}{k} \tr \left( ( \phi_{1:k}^\mathcal{T}  - \widehat UV\phi_{1:k}^\mathcal{H})( \phi_{1:k}^\mathcal{T}  - \widehat UV\phi_{1:k}^\mathcal{H})\tps\right)\nonumber\\
&= \frac{1}{k} \tr \left( {\phi_{1:k}^\mathcal{T}}\tps\phi_{1:k}^\mathcal{T} - 2{\phi_{1:k}^\mathcal{T}}\tps \widehat UV\phi_{1:k}^\mathcal{H} + {\phi_{1:k}^\mathcal{H}}\tps V\tps \widehat U\tps \widehat UV\phi_{1:k}^\mathcal{H} \right)\nonumber\\
\implies {\bf d}\mathcal{L} &= -2\text{tr}\left ( {\phi_{1:k}^\mathcal{H}}\tps{\bf d}V\tps \widehat U\tps \phi_{1:k}^\mathcal{T}\right ) + 2\text{tr}\left ( {\phi_{1:k}^\mathcal{H}}\tps {\bf d}V\tps \widehat U\tps\widehat U V\widehat \phi_{1:k}^\mathcal{H}\right )\nonumber\\
\implies {\bf d}\mathcal{L} &= -2\text{tr}\left ( {\bf d}V\tps \widehat U\tps \phi_{1:k}^\mathcal{T}{\phi_{1:k}^\mathcal{H}}\tps\right ) + 2\text{tr}\left (  {\bf d}V\tps \widehat U\tps\widehat U V\widehat \phi_{1:k}^\mathcal{H}{\phi_{1:k}^\mathcal{H}}\tps\right )\nonumber\\
\implies {\bf d}\mathcal{L} &= -2\text{tr}\left ( {\bf d}V\tps \widehat U\tps \widehat \Sigma_\mathcal{T,H}\right ) + 2\text{tr}\left (  {\bf d}V\tps \widehat U\tps\widehat U V\widehat \Sigma_\mathcal{H,H}\right )\nonumber\\
\implies \frac{{\bf d}\mathcal{L}}{{\bf d}V\tps} &= -2\text{tr}\left (  \widehat U\tps \widehat \Sigma_\mathcal{T,H}\right ) + 2\text{tr}\left ( \widehat U\tps\widehat U V\widehat \Sigma_\mathcal{H,H}\right )\nonumber\\
\implies 0 &=   -\widehat U\tps \widehat \Sigma_\mathcal{T,H} + \widehat U\tps\widehat U V\widehat \Sigma_\mathcal{H,H}\nonumber\\
\implies \widehat V &= (\widehat U\tps\widehat U)^{-1}\widehat U\tps \widehat \Sigma_\mathcal{T,H}( \widehat \Sigma_\mathcal{H,H})^{-1} \nonumber\\
&= \widehat U\tps  \widehat \Sigma_\mathcal{T,H}( \widehat \Sigma_\mathcal{H,H})^{-1} 
\end{align*}

\section{Experimental Results}
\label{sec:AppendExp}
\subsection{RR-POMDP}
The RR-POMDP parameters are:\\

 [$m=4$ hidden states, $n=2$ observations, $k=3$ transition matrix rank].
\begin{align*}
T^\pi = \left [ \begin{array}{cccc}
     0.7829    &0.1036    &0.0399    &0.0736\\
     0.1036    &0.4237    &0.4262    &0.0465\\
     0.0399    &0.4262    &0.4380    &0.0959\\
     0.0736    &0.0465    &0.0959    &0.7840\end{array} \right] \ \ \ \
O = \left [\begin{array}{cccc}
1& 0 & 1 & 0\\
0  & 1 & 0 & 1
  \end{array}  \right]
\end{align*}
The discount factor is $\gamma = 0.9$.
\subsection{Pricing a financial derivative}
\paragraph{Basis functions}The fist 16 are the basis functions suggested by Van Roy; for full description and justification see~\cite{Tsitsiklis97,Choi2006}. The first functions consist of a constant,  the reward, the minimal and maximal returns, and how long ago they occurred:
\begin{align*}
\phi_1(x) &= 1\\
\phi_2(x) &= G(x)\\
\phi_3(x) &= \min_{i=1,\hdots,100}x(i) - 1\\
\phi_4(x) &= \max_{i=1,\hdots,100}x(i) - 1\\
\phi_5(x) &= \argmin_{i=1,\hdots,100}x(i) - 1\\
\phi_6(x) &= \argmax_{i=1,\hdots,100}x(i) - 1
\end{align*}
The next set of basis functions summarize the characteristics of the basic shape of the 100 day sample path. They are the inner product of the path with the first four Legendre polynomial degrees. Let $j = i/50 -1$.
\begin{align*}
\phi_7(x) &= \frac{1}{100} \sum_{i=1}^{100} \frac{x(i) -1}{\sqrt{2}}\\
\phi_8(x) &= \frac{1}{100} \sum_{i=1}^{100}x(i)\sqrt{\frac{3}{2}}j\\
\phi_9(x) &= \frac{1}{100} \sum_{i=1}^{100}x(i)\sqrt{\frac{5}{2}}\left( \frac{3j^2 - 1}{2} \right)\\
\phi_{10}(x) &= \frac{1}{100} \sum_{i=1}^{100}x(i)\sqrt{\frac{7}{2}}\left( \frac{5j^3 - 3j}{2} \right)
\end{align*}
Nonlinear combinations of basis functions:
\begin{align*}
\phi_{11}(x) &= \phi_2(x)\phi_3(x)\\
\phi_{12}(x) &=  \phi_2(x)\phi_4(x)\\
\phi_{13}(x) &=  \phi_2(x)\phi_7(x)\\
\phi_{14}(x) &= \phi_2(x)\phi_8(x)\\
\phi_{15}(x) &=  \phi_2(x)\phi_9(x)\\
\phi_{16}(x) &=  \phi_2(x)\phi_{10}(x)
\end{align*}
In order to improve our results, we added a large number of additional basis functions to these hand-picked 16. PSTD will compress these features for us, so we can use as many additional basis functions as we would like. First we defined 4 additional basis functions consisting of the inner products of the 100 day sample path with the 5th and 6th Legende polynomials and we added the corresponding nonlinear combinations of basis functions:
\begin{align*}
\phi_{17}(x) &= \frac{1}{100} \sum_{i=1}^{100}x(i)\sqrt{\frac{9}{2}}\left( \frac{35j^4 - 30x^2 + 3}{8} \right)\\
\phi_{18}(x) &= \frac{1}{100} \sum_{i=1}^{100}x(i)\sqrt{\frac{11}{2}}\left( \frac{63j^5 - 70j^3 + 15j}{8} \right)\\
\phi_{19}(x) &= \phi_2(x)\phi_{17}(x)\\
\phi_{20}(x) &=  \phi_2(x)\phi_{18}(x)
\end{align*}
Finally we added the the entire sample path and the squared sample path:
\begin{align*}
\phi_{21:120} &= x_{1:100}\\
\phi_{121:220} &= x_{1:100}^2
\end{align*}

\end{document}